\documentclass[pdflatex,sn-vancouver-num]{sn-jnl}%

\usepackage{manyfoot}
\usepackage{xcolor}

\usepackage{graphicx}
\usepackage{multirow}
\usepackage{makecell}
\usepackage{amsmath,amssymb,amsfonts}
\usepackage{booktabs}
\usepackage{textcomp}
\usepackage{xspace}
\usepackage{colortbl}

\newcommand{\mus}{\textmu US\xspace}

\begin{document}

\title[ProtoCouncil]{ProtoCouncil: Multi-Encoder Anatomical Prototypes for Prostate Cancer Detection in Micro-Ultrasound}

\author*[1,3]{\fnm{Mohammad Mahdi} \sur{Abootorabi}} \email{mahdi.abootorabi@ece.ubc.ca} 
\author[2,3]{\fnm{Nooshin} \sur{Maghsoodi}}%
\author[1]{\fnm{Armin} \sur{Saadat}}%
\author[1,3]{\fnm{Obed} \sur{Dzikunu}}%
\author[2,3]{\fnm{Zhuoxin} \sur{Guo}}%
\author[4]{\fnm{Hannes} \sur{Cash}}%
\author[5]{\fnm{Ferdinand} \sur{Luger}}%
\author[6]{\fnm{Adam} \sur{Kinnaird}}%
\author[7]{\fnm{Brian} \sur{Wodlinger}}%
\author[2,3]{\fnm{Parvin} \sur{Mousavi}}
\equalcont{Co-senior authorship.}
\author[1]{\fnm{Purang} \sur{Abolmaesumi}}
\equalcont{Co-senior authorship. Code: \href{https://github.com/DeepRCL/ProtoCouncil}{github.com/DeepRCL/ProtoCouncil}}

\affil[1]{\orgname{The University of British Columbia}, \orgaddress{\city{Vancouver}, \state{BC}, \country{Canada}}}
\affil[2]{\orgname{Queen's University}, \orgaddress{\city{Kingston}, \state{ON}, \country{Canada}}}
\affil[3]{\orgname{Vector Institute}, \orgaddress{\city{Toronto}, \state{ON}, \country{Canada}}}
\affil[4]{\orgname{ProUro}, \orgaddress{\city{Berlin}, \country{Germany}}}
\affil[5]{\orgname{Ordensklinikum Linz}, \orgaddress{\city{Linz}, \country{Austria}}}
\affil[6]{\orgname{University of Alberta}, \orgaddress{\city{Edmonton}, \state{AB}, \country{Canada}}}
\affil[7]{\orgname{Exact Imaging}, \orgaddress{\city{Markham}, \state{ON}, \country{Canada}}}

\abstract{
\textbf{Purpose:} Micro-ultrasound (\mus) is a promising modality for prostate cancer (PCa) detection, but identifying suspicious tissue depends heavily on clinical experience. Models trained on sparse, noisy core-level pathology offer little insight into their decisions and can fail under domain shift.
Clinical use needs a traceable decision process and a flag for uncertain predictions.

\textbf{Methods:} We introduce ProtoCouncil, in which three foundation encoders (MedSAM, DINOv3 and V-JEPA 2.1) learn case-based prototypes at three anatomical levels: whole gland, needle track and gland sectors. Prototypes are projected onto real training-core regions, and label-free fusion keeps every class logit as a bias plus an exact sum of prototype similarity contributions. Encoder disagreement triggers selective prediction.

\textbf{Results:} In five-fold patient-level cross-validation (6,588 cores, 693 patients), ProtoCouncil reaches 82.9\% AUROC for clinically significant PCa (csPCa, Grade Group $\geq$3), 4.3 points above the best previously published method. In zero-shot evaluation on two external cohorts, ProtoCouncil achieves higher all-cancer and csPCa AUROC than all previous methods. Nearest same-class exemplars match test-core tumor involvement better than random same-class cases. Externally, csPCa accuracy falls from 79\% when encoders agree to 27\% when two oppose the fused call. Abstaining on disagreement raises balanced accuracy by 7 and 15 points, while retaining 58\% and 54\% of the cores.

\textbf{Conclusion:} 
ProtoCouncil grounds core-level predictions in exact similarity-based evidence from real, anatomically indexed training cases, which makes its reasoning interpretable. It achieves the highest csPCa AUROC across three cohorts and identifies unreliable predictions through encoder disagreement.
}


\maketitle

\section{Introduction}\label{sec:intro}

Prostate cancer (PCa) remains a leading cause of cancer mortality in men. MRI-targeted biopsy improves the detection of clinically significant PCa (csPCa), but the cost and infrastructure requirements of MRI limit its accessibility. High-resolution micro-ultrasound (\mus) provides real-time prostate imaging at 29~MHz, and \mus-targeted biopsy has been shown to be non-inferior to MRI-targeted biopsy~\cite{kinnaird2025microultrasonography}. 
However, interpretation of \mus remains highly dependent on clinical experience. The Prostate Risk Identification using Micro-Ultrasound (PRI-MUS) protocol~\cite{ghai2016primus} defines a visual vocabulary for suspicious tissue, but its application varies across readers, which motivates machine-learning tools to
support clinical assessment.

A key challenge is the limited supervision in this setting.
Biopsy histopathology provides an International Society of
Urological Pathology (ISUP) Grade Group (GG) and cancer involvement (the percentage of the sampled tissue
occupied by cancer) for each core.
However, these labels do not indicate where cancer is located within the corresponding \mus image. Supervision is therefore sparse along the needle track, noisy because small tumors may be missed, and strongly imbalanced because benign cores dominate. Recent work has progressed from multiple-instance transformers on needle patches to foundation models (FMs) adapted to these labels with clinical prompts~\cite{wilson2024prostnfound,wilson2025prostnfoundplus}, reinforcement-learning attention~\cite{abootorabi2026prostrl}, histopathology distillation~\cite{elghareb2026guideus}, and test-time adaptation~\cite{dzikunu2026ant}.
However, these models provide limited insight into the evidence behind their predictions. Predictive entropy can also be low even when a prediction is incorrect. These limitations motivate an approach that links predictions to real training examples and uses an estimator to flag potentially unreliable decisions. 

Prototype networks offer a different kind of transparency. An input is classified by its similarity to prototypes that are projected onto real training examples, so the decision process can be traced to training data~\cite{chen2019protopnet,ma2024protovit}.
Prototype models have also been applied to ultrasound~\cite{gu2025protoasnet,vaseli2025hiprotonet}, but they typically match prototypes to small learned image or video regions using a single encoder. Applying these approaches to prostate cancer detection in \mus images raises two challenges.
First, a biopsy label describes tissue sampled along a needle track within an anatomically structured gland, rather than a localized image region. Prototypes matched to arbitrary image patches therefore lack an explicit anatomical reference.
Second, limited labeled prostate \mus data motivate adapting pretrained FMs, but a single adapted FM may capture useful yet incomplete features. FMs pretrained on medical images, natural images, and videos may capture complementary features, so combining their case-based predictions could improve detection, and their disagreement could help identify unreliable predictions.

\begin{figure}[t]
\centering
\includegraphics[width=\textwidth]{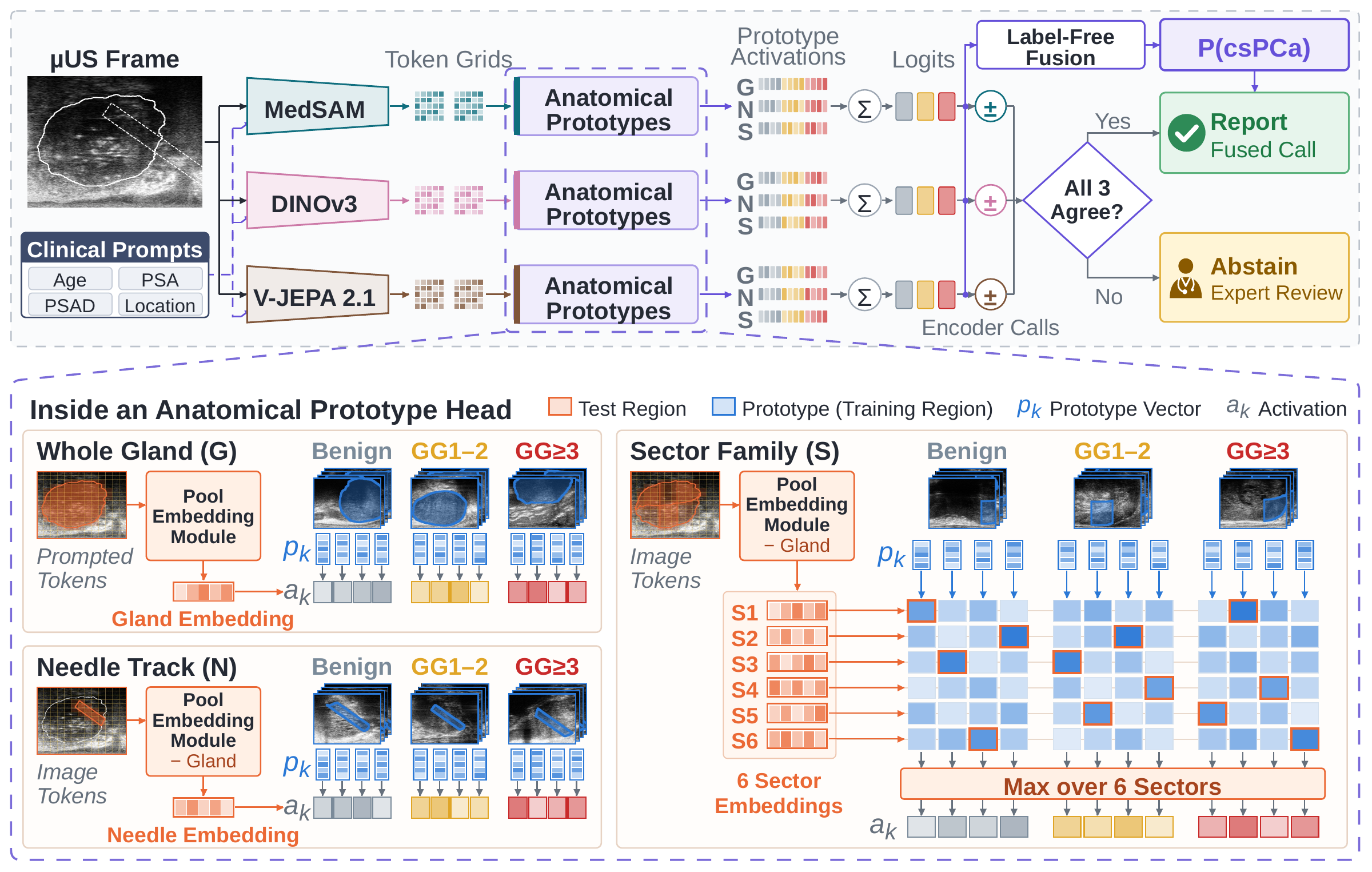}
\caption{
ProtoCouncil overview. Three encoders feed anatomical prototype heads whose logits are fused without labels. The model reports the fused prediction only when all three encoders agree; otherwise, it abstains. Bottom: each head compares pooled gland, needle-track, and gland-relative sector embeddings with prototypes from training regions.
}\label{fig:overview}
\end{figure}

\noindent
\textbf{Contributions.} In this work, (i) we introduce ProtoCouncil (Fig.~\ref{fig:overview}) for core-level PCa detection, with a decision process traceable to training data. We learn class-specific prototypes for the whole gland, needle track, and six gland sectors, and project them onto real training regions, which makes the model's decisions interpretable;
(ii) we learn these prototypes using MedSAM~\cite{ma2024medsam}, DINOv3~\cite{simeoni2026dinov} and V-JEPA~2.1~\cite{murlabadia2026vjepa21} token embeddings and fuse them while preserving each encoder's prototype contributions.
Across three cohorts (8,606 cores, 846 patients), ProtoCouncil achieves higher csPCa AUROC point estimates than previous methods;
(iii) we show that encoder disagreement identifies less reliable predictions: external csPCa accuracy falls from 78--79\% when no encoder opposes the fused prediction to 27\% when two encoders oppose it.

\section{Methods}\label{sec:methods}

\subsection{Data}\label{sec:data}
We use three \mus cohorts acquired with the ExactVu system (Exact Imaging, Markham, Canada) in approved clinical trials with informed consent (Table~\ref{tab:cohorts}). Cohort A comes from NCT02079025, while B and C come from OPTIMUM~\cite{kinnaird2025microultrasonography}. Each core has a sagittal B-mode frame acquired before needle firing, resized to $256\times256$, with prostate and needle-trace masks. 
Following~\cite{wilson2025prostnfoundplus,abootorabi2026prostrl}, we define all-cancer as GG$\geq$1 and csPCa as GG$\geq$3, and assign patients the highest GG among their cores. Clinical prompts include age, PSA, PSA density, and core location. PSA density is unavailable in B and C and is imputed with the corresponding cohort-A training mean. Cohort A follows patient-level, center-stratified five-fold cross-validation~\cite{abootorabi2026prostrl}.
We then evaluate all models zero-shot on cohorts B and C.

\begin{table}[b]
\centering
\caption{Characteristics of development cohort A and external cohorts B and C.
Pathology categories show core counts (percentages).
csPCa: GG$\geq$3; sys./targ.: systematic/targeted;
TR/TP: transrectal/transperineal.}
\label{tab:cohorts}
\setlength{\tabcolsep}{2.5pt}
\begin{tabular}{@{}lllrrrrrr@{}}
\toprule
Cohort & Sites & Biopsy & Patients & Cores & Benign & GG1--2
& GG$\geq$3 & \makecell[r]{csPCa\\patients} \\
\midrule
A & 5 & Sys., TR & 693 & 6,588
& 5,709 (86.7\%) & 480 (7.3\%) & 399 (6.1\%) & 117 \\
B & 2 & Sys.\,+\,targ., TR & 111 & 1,460
& 982 (67.3\%) & 364 (24.9\%) & 114 (7.8\%) & 28 \\
C & 1 & Sys.\,+\,targ., TP & 42 & 558
& 433 (77.6\%) & 105 (18.8\%) & 20 (3.6\%) & 6 \\
\bottomrule
\end{tabular}
\end{table}

\subsection{Anatomically Aware Prototype Heads}\label{sec:heads}

\textbf{Encoders.} ProtoCouncil builds on three FMs pretrained on different data and tasks: MedSAM ViT-B~\cite{ma2024medsam} for medical image segmentation, DINOv3 ViT-L/16~\cite{simeoni2026dinov} through self-supervised learning on natural images, and V-JEPA~2.1 ViT-L/16~\cite{murlabadia2026vjepa21} through self-supervised learning on images and videos. These different pretraining settings can provide complementary representations of \mus tissue appearance.
Each FM is fine-tuned on the training folds as a prompt-conditioned cancer detector~\cite{wilson2025prostnfoundplus,abootorabi2026prostrl}, with image tokens attending to clinical prompt tokens in the decoder. We then freeze each detector and extract two $16\times16$ token grids: image features from the encoder (enc) and prompt-conditioned contextual features (ctx) from the decoder. 

\noindent
\textbf{Anatomical regions.} 
In natural images, prototype matches can correspond to recognizable object parts, such as a bird's head~\cite{chen2019protopnet}. In \mus, patch matches can reflect local tissue texture without a consistent anatomical reference across cores. Finer anatomy cannot replace these patches, since no zonal or lesion segmentation of these images is available for training, and a biopsy label does not say where cancer lies along the needle. 
We therefore describe each frame by three families of anatomical regions derived from its masks: the whole gland $G$, the needle track $N$ (needle trace $\cap$ gland) and six sectors that split the gland into posterior and anterior halves by per-column relative depth and into three equal-area thirds along its horizontal axis.

\noindent
For each region $R$, the descriptor $d(R)$ concatenates the overlap-weighted mean of the tokens in region $R$ with their channel-wise maximum over tokens at least half inside $R$ (to preserve focal patterns that averaging over a large region would dilute). To retain image-level context while emphasizing regional differences, we represent the gland by the prompt-conditioned descriptor $d^{\mathrm{ctx}}(G)$ and the needle track and sectors by $d^{\mathrm{enc}}(R)-d^{\mathrm{enc}}(G)$, which use image features only. Regional prototypes therefore compare features relative to each core's own gland. Two projectors (LayerNorm, Linear to 128, GELU, Linear to 64) map descriptors to unit-norm embeddings $u(R)$: one for the gland, and one shared by the needle track and sectors, whose embeddings therefore lie in a common space.

\noindent
\textbf{Prototype matching and class scores.} Following prototype networks such as ProtoPNet~\cite{chen2019protopnet}, each encoder learns class-specific prototypes for the gland, needle-track and sector families, with four prototypes per family and class (benign, GG1--2 and GG$\geq$3), i.e., 36 per encoder. Prototypes $p_k$ and region embeddings $u(R)$ are unit-norm vectors, so $u(R)^{\top}p_k$ is their cosine similarity, which we rescale to $[0,1]$ as $s_k(R)=\tfrac12(1+u(R)^{\top}p_k)$. We score a prototype by its best-matching region: the activation $a_k$ is the highest similarity over the regions of its family (the gland, the needle track or one of the six sectors), and this region is recorded for the explanations. A final linear layer maps prototype activations to class logits and gives:
\begin{equation}
z_c = b_c + \textstyle\sum_{k\in\mathcal{P}_c} w_k\, a_k, \qquad w_k=\operatorname{softplus}(\rho_k)\geq 0,
\label{eq:readout}
\end{equation}
where $\mathcal{P}_c$ contains the prototypes of class $c$ and $b_c$ is a bias. Each prototype contributes only to its own class logit, and the rescaled similarities and non-negative weights make every contribution non-negative. Each logit is thus a bias plus a sum of weighted similarities. A softmax over the three logits gives $P(\text{cancer})=1-P(\text{benign})$ and $P(\text{csPCa})=P(\text{GG}{\geq}3)$. 
We train with a class-weighted cross-entropy $\mathcal{L}_{\mathrm{CE}}$ and with cluster and separation terms $\mathcal{L}_{\mathrm{clst}}$ and $\mathcal{L}_{\mathrm{sep}}$ adapted from previous prototype-based models, which encourage a close match between one of a core's regions and a prototype of its class and penalize strong matches to prototypes of other classes. A diversity penalty $\mathcal{L}_{\mathrm{div}}$ discourages prototypes of the same class and family from becoming too similar, giving the total loss $\mathcal{L}=\mathcal{L}_{\mathrm{CE}}+\lambda_{\mathrm{clst}}\mathcal{L}_{\mathrm{clst}}+\lambda_{\mathrm{sep}}\mathcal{L}_{\mathrm{sep}}+\lambda_{\mathrm{div}}\mathcal{L}_{\mathrm{div}}$.

\noindent
\textbf{Family balancing.} The gland descriptor summarizes the whole prostate together with the clinical context, so a head can come to rely mostly on the gland. To reduce this reliance, we randomly replace each core's gland activations with their stop-gradient batch mean during training (probability 0.5). This encourages the model to also draw evidence
from the needle and sector prototypes.

\noindent
\textbf{Projection.} After training, a Hungarian assignment replaces the encoder's 36 prototypes with embeddings of real training regions from 36 distinct patients, which maximizes total cosine similarity. Candidates are training cores of the prototype's class (at least 40\% involvement for the cancer classes) with enough region support. Projectors and prototypes are then frozen, and only the final
classification layer is refit. Consequently, every prototype equals the embedding of an identifiable training region.

\noindent
\textbf{Encoder score fusion.}\label{sec:fusion}
We combine the encoder-specific predictions through a weighted average of their class logits. To account for differences in logit scale across encoders, we assign weights inversely proportional to their logit
standard deviations:
\begin{equation}
z_c = \sum_e \alpha_e z_c^{(e)}, \qquad
\alpha_e =
\frac{\sigma_e^{-1}}{\sum_{e'} \sigma_{e'}^{-1}}.
\label{eq:fusion}
\end{equation}
Here, $c$ denotes the diagnostic class and $e$ and $e'$ index the encoders. The encoder-specific logit $z_c^{(e)}$ contributes to the fused logit $z_c$ with weight $\alpha_e$. 
The standard deviation $\sigma_e$ is computed on the validation fold after subtracting the mean across classes from each core's logits. 
The weights are estimated without labels and remain fixed at inference. 
The weighted combination preserves the additive structure of each encoder's class scores (Eq.~\ref{eq:readout}): the fused logit remains a bias plus an exact sum of weighted prototype similarities. The combined prediction therefore remains traceable to the training regions represented by the prototypes.

\noindent
\textbf{Encoder disagreement and selective prediction.}
\label{sec:selective}
Each encoder and the fused model predict csPCa when $p=P(\text{GG}{\geq}3)$ reaches a model-specific threshold $t$. Thresholds are set to 80\% specificity on development validation and applied unchanged to other external datasets. We describe disagreement by the number of encoders whose predictions differ from the fused prediction, $O\in\{0,\dots,3\}$. For selective prediction, we retain the fused prediction only when all three encoders agree; otherwise, the model abstains.
We can also perform the selection at fixed coverage, defined as the proportion of cores retained. 
As a baseline, a confidence gate retains the cores farthest from the threshold, those with the largest $|\operatorname{logit}(p)-\operatorname{logit}(t)|$. 
Because coverage can be chosen freely for this gate, we evaluate it at the coverage of the disagreement gate.

\subsection{Implementation and Evaluation}
\label{sec:implementation}
Prototype heads are trained on cached tokens using AdamW, with learning rates of $10^{-3}$ for the projectors and prototypes and $10^{-2}$ for the final classification layer, and a batch size of 96. Training runs for up to 40 epochs, with early stopping based on the mean validation AUROC for all-cancer and csPCa.
Baseline methods are trained on the same patient folds. MedSAM-UNETR~\cite{alzate2023samunetr}, MedSAM-FT~\cite{ma2024medsam}, MicroSegNet~\cite{jiang2024microsegnet} and ProstNFound~\cite{wilson2024prostnfound} are scored by their mean heatmap value within the needle--prostate intersection, whereas ProstNFound+~\cite{wilson2025prostnfoundplus}, GUIDE-US~\cite{elghareb2026guideus}, Prost-RL~\cite{abootorabi2026prostrl} and DINO-FT use their classifier outputs.
DINO-FT is a prompt-conditioned DINOv3 detector that we fine-tuned on the same folds.
For each fold, we average the predicted probabilities of two head seeds; displayed explanations describe a single fold and head seed.

We report AUROC for all-cancer, csPCa, high-involvement cancer and sensitivity at 60\% or 80\% specificity for both endpoints. 
Patient-level csPCa scores are obtained by averaging core scores. Results are reported as mean $\pm$ SD across the five fold models, evaluated on their respective validation folds and on each complete
external test cohort.
Method comparisons use the mean of paired differences across fold models. Confidence intervals and $p$-values are calculated using a paired patient-level bootstrap stratified by center (10,000 resamples), with clustered DeLong tests for AUROC as a sensitivity analysis.
We corrected $p$-values with the Holm method for comparisons.

\section{Results and Discussion}\label{sec:results}

\begin{table}[!t]
\centering
\caption{Prostate cancer detection in micro-ultrasound (\%, mean$_{\pm\mathrm{SD}}$ over the five fold models). All: all cancer (GG$\geq$1); High-inv.: benign vs.\ cancer cores with $>$40\% involvement; csPCa: clinically significant cancer (GG$\geq$3). Best and second-best are shown in \textbf{bold} and \underline{underlined}.}\label{tab:detection}
\newcommand{\ms}[2]{$#1_{\pm#2}$}
\newcommand{\bms}[2]{$\mathbf{#1}_{\pm#2}$}
\newcommand{\ums}[2]{$\underline{#1}_{\pm#2}$}
\setlength{\tabcolsep}{2.2pt}
\begin{tabular}{@{}lccccccc@{}}
\toprule
& \multicolumn{2}{c}{All} & High-inv. & \multicolumn{3}{c}{csPCa} & Patient \\
\cmidrule(lr){2-3}\cmidrule(lr){4-4}\cmidrule(lr){5-7}\cmidrule(lr){8-8}
Method & AUROC & Sens@80 & AUROC & AUROC & Sens@60 & Sens@80 & AUROC \\
\midrule
\multicolumn{8}{@{}l}{\textit{Cohort A (development, five-fold cross-validation)}} \\
MedSAM-UNETR~\cite{alzate2023samunetr} & \ms{70.1}{2.1} & \ms{49.4}{4.6} & \ms{79.7}{1.1} & \ms{71.4}{4.0} & \ms{71.4}{6.1} & \ms{50.7}{7.5} & \ms{67.4}{9.2} \\
MedSAM-FT~\cite{ma2024medsam} & \ms{72.2}{3.0} & \ms{49.9}{5.1} & \ms{80.0}{0.8} & \ms{72.5}{3.5} & \ms{73.4}{4.0} & \ms{48.4}{6.4} & \ms{67.0}{6.2} \\
MicroSegNet~\cite{jiang2024microsegnet} & \ms{76.3}{1.3} & \ms{57.7}{3.1} & \ms{83.2}{1.2} & \ms{77.6}{4.6} & \ms{78.4}{7.0} & \ms{62.7}{6.7} & \ums{73.7}{7.5} \\
ProstNFound~\cite{wilson2024prostnfound} & \ms{77.1}{2.9} & \ms{60.3}{4.6} & \ms{82.6}{1.9} & \ms{77.7}{4.8} & \ms{82.3}{7.1} & \ms{62.1}{9.2} & \ms{72.2}{3.3} \\
ProstNFound+~\cite{wilson2025prostnfoundplus} & \ms{77.3}{3.6} & \ms{60.9}{6.2} & \ms{82.3}{2.1} & \ms{78.0}{5.8} & \ms{80.9}{8.5} & \ms{58.6}{10.1} & \ms{72.0}{2.2} \\
Prost-RL (Supervised)~\cite{abootorabi2026prostrl} & \ums{78.3}{3.6} & \ums{62.4}{5.1} & \ms{83.7}{2.3} & \ms{78.1}{5.3} & \ums{84.1}{9.2} & \ms{57.0}{10.6} & \ms{70.6}{4.3} \\
Prost-RL~\cite{abootorabi2026prostrl} & \ms{78.1}{3.9} & \ms{62.2}{6.2} & \ms{83.3}{2.8} & \ms{78.6}{5.5} & \ms{83.9}{10.0} & \ms{63.3}{12.4} & \ms{70.5}{3.8} \\
GUIDE-US~\cite{elghareb2026guideus} & \ms{76.9}{3.0} & \ms{59.9}{4.7} & \ms{82.8}{2.8} & \ms{77.9}{5.2} & \ms{83.1}{8.4} & \ms{57.6}{10.0} & \ms{72.3}{3.2} \\
DINO-FT & \ms{78.0}{3.2} & \ms{60.9}{2.8} & \ums{85.2}{1.0} & \ums{79.1}{4.1} & \ms{79.9}{6.2} & \ums{63.5}{6.3} & \ms{73.1}{5.2} \\
\rowcolor[gray]{0.9}
ProtoCouncil (ours) & \bms{81.3}{2.8} & \bms{68.1}{3.9} & \bms{88.8}{1.0} & \bms{82.9}{4.3} & \bms{88.1}{3.0} & \bms{68.7}{10.8} & \bms{74.8}{5.7} \\
\midrule
\multicolumn{8}{@{}l}{\textit{Cohort B (external, zero-shot performance, two sites, transrectal)}} \\
MedSAM-UNETR~\cite{alzate2023samunetr} & \ms{68.1}{1.2} & \ms{46.4}{2.3} & \ms{76.3}{1.3} & \ms{72.9}{1.9} & \ms{74.7}{3.7} & \ms{52.6}{4.4} & \ms{66.9}{1.7} \\
MedSAM-FT~\cite{ma2024medsam} & \ms{67.0}{0.7} & \ms{43.5}{0.6} & \ms{75.8}{1.0} & \ms{72.6}{0.8} & \ms{73.7}{2.9} & \ms{49.1}{1.6} & \ms{59.9}{1.0} \\
MicroSegNet~\cite{jiang2024microsegnet} & \ms{69.4}{1.1} & \ms{47.7}{2.1} & \ms{78.0}{2.1} & \ms{74.6}{1.7} & \ms{76.0}{2.9} & \ms{58.1}{4.4} & \ms{65.7}{2.3} \\
ProstNFound~\cite{wilson2024prostnfound} & \ms{73.3}{1.9} & \ms{54.2}{4.2} & \ms{80.2}{2.1} & \ms{75.3}{2.1} & \ms{78.8}{3.4} & \ms{54.0}{5.5} & \ms{68.1}{1.7} \\
ProstNFound+~\cite{wilson2025prostnfoundplus} & \ums{74.5}{0.8} & \ums{54.6}{1.3} & \ms{82.7}{0.9} & \ms{78.0}{0.5} & \ms{81.9}{0.9} & \ms{61.6}{2.3} & \ms{68.6}{0.9} \\
Prost-RL (Supervised)~\cite{abootorabi2026prostrl} & \ms{72.5}{0.9} & \ms{51.5}{2.3} & \ms{80.3}{1.1} & \ms{75.6}{1.5} & \ms{75.8}{2.8} & \ms{53.9}{3.7} & \ms{68.4}{0.8} \\
Prost-RL~\cite{abootorabi2026prostrl} & \ms{72.5}{0.7} & \ms{52.0}{2.4} & \ms{80.1}{0.9} & \ms{75.4}{1.3} & \ms{77.0}{3.1} & \ms{52.6}{2.4} & \ms{68.8}{0.7} \\
GUIDE-US~\cite{elghareb2026guideus} & \ms{73.1}{0.6} & \ms{52.5}{1.2} & \ms{79.8}{1.4} & \ms{75.5}{2.2} & \ms{77.0}{4.5} & \ms{52.5}{5.1} & \ms{68.2}{1.3} \\
DINO-FT & \ms{73.1}{1.6} & \ms{53.9}{2.3} & \ums{82.9}{1.7} & \ums{79.5}{1.3} & \ums{82.8}{2.1} & \ums{65.1}{3.3} & \bms{72.2}{1.3} \\
\rowcolor[gray]{0.9}
ProtoCouncil (ours) & \bms{74.9}{0.7} & \bms{57.0}{2.3} & \bms{85.2}{0.5} & \bms{80.0}{1.3} & \bms{83.2}{1.2} & \bms{67.0}{3.9} & \ums{70.3}{2.0} \\
\midrule
\multicolumn{8}{@{}l}{\textit{Cohort C (external, zero-shot performance, one site, transperineal)}} \\
MedSAM-UNETR~\cite{alzate2023samunetr} & \ms{55.4}{1.0} & \ms{27.0}{3.9} & \ms{53.6}{1.5} & \ms{57.5}{3.0} & \ms{51.0}{5.8} & \ms{31.0}{8.6} & \ms{74.5}{2.3} \\
MedSAM-FT~\cite{ma2024medsam} & \ms{62.9}{1.3} & \bms{42.2}{1.7} & \ms{63.9}{1.4} & \ms{76.6}{1.3} & \ms{80.0}{4.5} & \ms{56.0}{6.6} & \ums{85.2}{2.2} \\
MicroSegNet~\cite{jiang2024microsegnet} & \ms{61.3}{0.9} & \ms{32.2}{3.0} & \bms{65.4}{1.9} & \ms{70.1}{3.4} & \ms{72.0}{5.1} & \ms{45.0}{6.3} & \ms{78.0}{5.7} \\
ProstNFound~\cite{wilson2024prostnfound} & \ms{59.3}{1.2} & \ms{30.9}{4.7} & \ms{57.2}{2.0} & \ms{62.9}{4.3} & \ms{56.0}{7.3} & \ms{22.0}{6.0} & \ms{69.3}{2.3} \\
ProstNFound+~\cite{wilson2025prostnfoundplus} & \ms{59.4}{2.3} & \ms{37.6}{1.2} & \ms{60.4}{2.7} & \ms{72.3}{1.7} & \ms{74.0}{2.0} & \ms{35.0}{10.5} & \ms{73.2}{4.2} \\
Prost-RL (Supervised)~\cite{abootorabi2026prostrl} & \ms{57.3}{1.2} & \ms{35.5}{2.5} & \ms{58.2}{1.0} & \ms{66.4}{2.3} & \ms{53.0}{14.7} & \ms{30.0}{4.5} & \ms{68.7}{1.4} \\
Prost-RL~\cite{abootorabi2026prostrl} & \ms{56.5}{1.5} & \ms{34.9}{1.2} & \ms{56.9}{1.8} & \ms{65.1}{2.9} & \ms{49.0}{12.4} & \ms{34.0}{2.0} & \ms{69.4}{0.7} \\
GUIDE-US~\cite{elghareb2026guideus} & \ms{57.1}{3.0} & \ms{30.2}{2.4} & \ms{55.9}{4.2} & \ms{61.3}{2.4} & \ms{39.0}{4.9} & \ms{25.0}{3.2} & \ms{61.5}{3.4} \\
DINO-FT & \ums{63.3}{2.9} & \ms{38.7}{2.3} & \ms{64.3}{2.9} & \ums{77.7}{3.2} & \ums{82.0}{8.1} & \ums{58.0}{9.3} & \ms{80.5}{2.6} \\
\rowcolor[gray]{0.9}
ProtoCouncil (ours) & \bms{63.6}{2.3} & \ums{40.0}{4.4} & \ums{64.6}{2.8} & \bms{79.5}{1.8} & \bms{84.0}{3.7} & \bms{60.0}{5.5} & \bms{86.5}{2.1} \\
\bottomrule
\end{tabular}
\end{table}

\subsection{Detection Performance}\label{sec:detection}
Main results are summarized in Table~\ref{tab:detection}. On cohort A, ProtoCouncil outperforms all previously published methods across different metrics. Its all-cancer AUROC (81.3\%) exceeds the strongest previous baseline by 3.0 percentage points and is significantly higher than that of every previous method. For csPCa, it reaches 82.9\%, 4.3 points above Prost-RL, the strongest previously published method (95\% CI [1.4, 7.3]). 
Externally, the margin narrows but ProtoCouncil keeps the highest csPCa AUROC, 80.0\% on B (vs.\ 78.0\% for ProstNFound+) and 79.5\% on C (vs.\ 76.6\% for MedSAM-FT). Its csPCa AUROC also changes little from the transrectal cohort B to the transperineal cohort C, whereas the previous prompt-conditioned methods, all built on MedSAM, lose 6--14 points. Since ProtoCouncil's own MedSAM prototype model also degrades on C (Sec.~\ref{sec:ablation}), this stability likely comes from combining encoders that transfer differently, although C includes only six patients with csPCa. Cohort C is also harder for all-cancer detection: GG1--2 makes up 84\% of its cancer cores, and no method exceeds 63.6\% all-cancer AUROC while csPCa AUROC stays near 80\%, which suggests that low-grade cancer is difficult to separate from benign tissue on \mus.
At the patient level, ProtoCouncil is highest on A and C but not on B (70.3\% vs.\ 72.2\% for DINO-FT), and none of these differences from the strongest methods is significant.

\begin{figure}[t]
\centering
\includegraphics[width=0.9\textwidth]{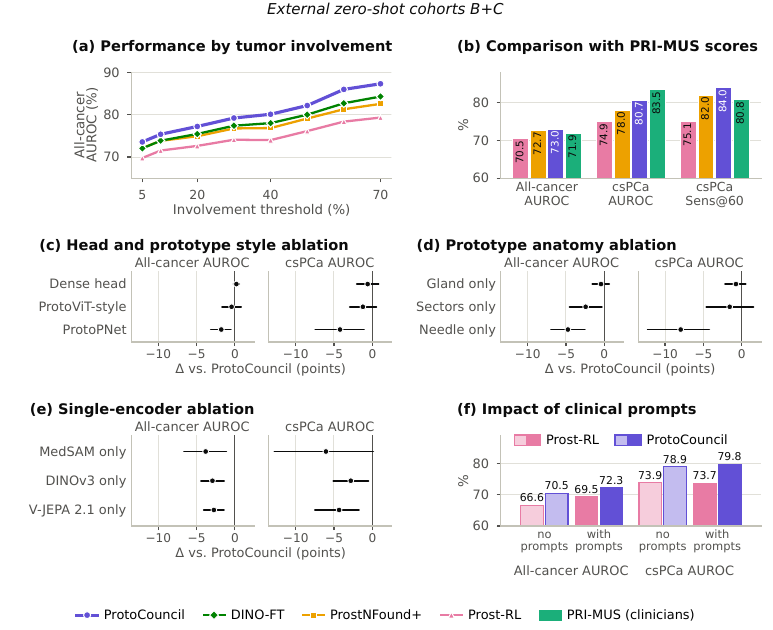}
\caption{
Results on the combined external zero-shot cohorts. \textbf{(a)} All-cancer AUROC by minimum cancer involvement. \textbf{(b)} Comparison with clinician-recorded
PRI-MUS scores (1,919 cores). \textbf{(c--e)} Head, anatomical-family
and encoder ablations: AUROC change relative to ProtoCouncil, with pointwise 95\% bootstrap CIs. \textbf{(f)} Effect of clinical prompts. 
}\label{fig:results}
\end{figure}

\noindent
Fig.~\ref{fig:results}\textbf{(a)} examines how detection performance changes with cancer involvement. Keeping benign cores fixed, all methods generally achieve higher AUROC as the minimum involvement required for cancer cores increases, which is consistent with greater difficulty in distinguishing cores containing small amounts of cancer from benign tissue. On the external cohorts, ProtoCouncil's
AUROC rises from 73.6\% at the 5\% threshold to 87.3\% at 70\% and remains above the other methods throughout. Its advantage persists across all thresholds and widens with involvement, from 1.4 to 4.8 points over ProstNFound+.

\begin{table}[!t]
\centering
\caption{
Selective csPCa prediction performance. Each gate abstains on the same share of cores as the disagreement gate, which abstains unless all three encoders agree (percentage of retained cores in parentheses). Balanced accuracy and csPCa AUROC of retained cores are reported.
}\label{tab:selective}
\setlength{\tabcolsep}{3pt}
\begin{tabular}{@{}lcccccc@{}}
\toprule
& \multicolumn{2}{c}{Cohort A (70\%)} & \multicolumn{2}{c}{Cohort B (58\%)} & \multicolumn{2}{c}{Cohort C (54\%)} \\
\cmidrule(lr){2-3}\cmidrule(lr){4-5}\cmidrule(lr){6-7}
& BAcc & AUROC & BAcc & AUROC & BAcc & AUROC \\
\midrule
\multicolumn{7}{@{}l}{\textit{No abstention}} \\
\quad Prost-RL~\cite{abootorabi2026prostrl} & 71.2 & 78.6 & 68.0 & 75.4 & 54.3 & 65.1 \\
\quad ProtoCouncil (ours) & 74.1 & 82.9 & 72.4 & 80.0 & 70.3 & 79.5 \\
\midrule
\multicolumn{7}{@{}l}{\textit{Abstention}} \\
\quad Prost-RL, confidence & 77.6 & 81.9 & 76.2 & 81.4 & 70.4 & 81.9 \\
\quad ProtoCouncil, confidence & 81.0 & 85.9 & 78.6 & 83.1 & 83.1 & 88.1 \\
\quad ProtoCouncil, disagreement (ours) & 80.2 & 86.6 & 79.0 & 84.8 & 84.8 & 90.9 \\
\bottomrule
\end{tabular}
\end{table}

\subsection{Encoder Disagreement and Selective Prediction}\label{sec:selective-results}
Encoder disagreement flags unreliable predictions. On external cohorts, the fused prediction is correct in 78.6\% of cores where all three encoders agree with it (56.9\% of the cores). Accuracy drops to 59.7\% when one encoder disagrees (34.5\%) and to only 27.0\% when two disagree (8.5\%). Abstaining whenever any encoder disagrees therefore retains 54--70\% of the cores and raises balanced accuracy by 6--15 points across the three cohorts (Table~\ref{tab:selective}).
On the external cohorts, at the same retained share this gate exceeds the confidence gate, with 1.7–2.8-point higher csPCa AUROC on the retained cores. It also outperforms the confidence gate of Prost-RL on every cohort. Unlike a confidence gate, it requires no retention level to be chosen: it abstains more often on the external cohorts, where the model is less accurate as it is zero-shot, and each abstention also points to the encoders whose prototype evidence conflicts.

\subsection{Case-based Evidence and Qualitative Analysis}\label{sec:evidence}

\begin{figure}[t]
\centering
\includegraphics[width=\textwidth]{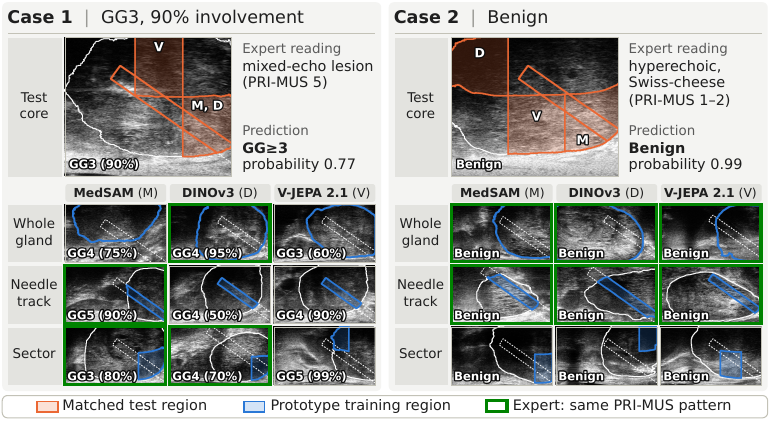}
\caption{
Qualitative examples showing how clinicians can trace ProtoCouncil predictions to labeled training examples and compare the matched tissue patterns. 
For each encoder and anatomical family, each cell shows a prototype that strongly contributes to the prediction.
Blue and orange mark prototype training regions and matched test regions, respectively. Letters (M, D, V) identify the encoder associated with each matched sector prototype. Labels report whole-core Grade Group and cancer involvement. Green frames mark pairs that an expert judged to show the same PRI-MUS pattern.
}
\label{fig:qualitative}
\end{figure}

The PRI-MUS protocol provides both a clinical reference for detection and a vocabulary for describing suspicious tissue. On the external cores with a recorded PRI-MUS score (Fig.~\ref{fig:results}\textbf{b}), ProtoCouncil outperforms the previous methods on all three metrics, and its all-cancer AUROC and csPCa sensitivity at 60\% specificity also exceed the PRI-MUS scores assigned by the biopsy clinicians. 
PRI-MUS retains a higher csPCa AUROC and a higher patient-level AUROC (81.4\% vs.\ 74.0\%). This gap reflects the core-level design of ProtoCouncil, which scores each core independently, whereas clinicians assess the whole gland.

As a complementary analysis, we examined whether prototype matches also reflect tumor involvement. For cancer cores with recorded
involvement, we compared the absolute involvement difference to the most similar exemplar in each anatomical family with the
mean difference to same-class exemplars. Gland matches reduced this difference by 6.1 and 6.8 percentage points on B and C, respectively. These exploratory findings suggest a coarse
correspondence between prototype similarity and tumor burden. Fig.~\ref{fig:qualitative} shows examples for a cancer and a benign core, with one prototype per encoder and anatomical family. Highlighted pairs share PRI-MUS patterns identified by an expert, while the prototype contributions make their role in the model's decision transparent.

\subsection{Ablation Studies}\label{sec:ablation}

Fig.~\ref{fig:results}\textbf{(c--f)} ablates the design choices of ProtoCouncil on the external cohorts.

\noindent
\textbf{Prediction heads (c).} A dense (non-prototype) head on the same descriptors is within 0.6 points of ProtoCouncil on both endpoints, so restricting the decision to prototype similarities costs no accuracy.
To assess the value of anatomical matching, we compare two alternative designs using the same frozen features, prototype count and fusion: ProtoPNet matches each prototype to a local
patch, whereas ProtoViT-style prototypes match groups of neighboring image tokens. ProtoPNet has significantly lower all-cancer and csPCa AUROC. ProtoViT-style matching also gives lower point estimates. 
These results support anatomical matching over classic patch prototypes, and each anatomical match also has a consistent reference that can be inspected.

\noindent
\textbf{Anatomical families and encoders (d, e).} The gland carries most of the evidence: a gland-only head is within 0.8 points of the full model, whereas needle-only and sector-only heads are 1.6--8.0 points lower. The regional families thus mainly add case-level evidence rather than detection performance. No single encoder is best on every cohort: MedSAM has the highest csPCa AUROC on A but drops to 59.8\% on C, where DINOv3 transfers best, and every single-encoder model has a lower csPCa AUROC than the council. The encoders also disagree on 30--46\% of the cores, which is what makes their disagreement a useful abstention signal (Sec.~\ref{sec:selective-results}).

\noindent
\textbf{Clinical prompts (f).} Consistent with previous studies~\cite{abootorabi2026prostrl}, clinical prompts help: removing them from every stage lowers ProtoCouncil's AUROC, most clearly on A (by 3.8 and 4.6 points). 
However, without prompts, ProtoCouncil still outperforms Prost-RL without prompts on the external cohorts (Holm $p=0.003$ for all-cancer and $0.012$ for csPCa AUROC), and even has a higher mean AUROC than Prost-RL with prompts.
Its advantage therefore does not depend on the clinical metadata alone.

\noindent
\textbf{Projection, capacity and fusion.} Projecting prototypes onto real training regions costs no accuracy: learned (unprojected) prototypes, 9 or 15 or 18 prototypes per class instead of 12, and training without family balancing all change csPCa AUROC by at most 0.5 points. The fusion weights do matter. Our weights improve csPCa AUROC over equal weights by 1.0 point on A.

\section{Conclusion}\label{sec:discussion}
We show that core-level PCa detection in \mus can achieve strong performance with a decision process that is traceable to training data. Matching anatomically defined regions to prototypes projected onto real training regions makes every class logit a bias plus a sum of weighted similarities to identifiable training cases (``this looks like that'' type of reasoning \cite{chen2019protopnet}), with accuracy comparable to a dense head and higher than classic patch prototypes. Using three foundation encoders addresses differences in how representations transfer across cohorts: no single encoder performed best on every cohort, whereas the council achieved the highest csPCa AUROC point estimates on all three. Their disagreement adds a second benefit: predictions opposed by more encoders were much less often correct, which gives an abstention rule that requires no retention level to be chosen and shows which encoders conflict. 
In future work, we will investigate whether anatomically indexed prototypes can serve as context for vision-language foundation models that combine information across the frames of a sweep and the cores of a patient, which may improve sweep- and patient-level detection. Whether presenting supporting and opposing training regions helps clinicians remains a question for future studies.

\noindent
\textbf{Acknowledgments.}
This work was supported in part by the Canadian Institutes of Health Research (CIHR), the Natural Sciences and Engineering Research Council of Canada (NSERC), Vector Institute, and through computational resources and services provided by Advanced Research Computing at the University of British Columbia. P. Mousavi is supported in part by Canada CIFAR AI Chair and Canada Research Chair.

\noindent\textbf{Disclosure of Interests.}
B.~Wodlinger is Vice President of Clinical and Engineering of Exact Imaging, the manufacturer of the micro-ultrasound system used in this study. The other authors have no competing interests to declare.

\bibliography{references}

\end{document}